\documentclass{article}

\usepackage{arxiv}
\usepackage{caption}
\usepackage{graphicx}
\usepackage[utf8]{inputenc} 
\usepackage[T1]{fontenc}    
\usepackage{hyperref}       
\usepackage{url}            
\usepackage{booktabs}       
\usepackage{amsfonts}       
\usepackage{nicefrac}       
\usepackage{microtype}      
\usepackage{lipsum}
\usepackage{listings}

\title{Classification of pedagogical content using conventional machine learning and deep learning model}

\author{
  Vedat Apuk, Krenare Pireva Nuçi  \\ 
  Department of Computer Science and Engineering\\
  University for Business and Technology\\
  10000 Prishine, Kosovo \\
  \texttt{va40172@ubt-uni.net, krenare.pireva@ubt-uni.net} \\

}

\begin{document}
\maketitle

\begin{abstract}
The advent of the Internet and a large number of digital technologies has brought with it many different challenges. A large amount of data is found on the web, which in most cases is unstructured and unorganized, and this contributes to the fact that the use and manipulation of this data is quite a difficult process. Due to this fact, the usage of different machine and deep learning techniques for Text Classification has gained its importance, which improved this discipline and made it more interesting for scientists and researchers for further study. 
This paper aims to classify the pedagogically content using two different models, the K-Nearest Neighbor (KNN) from the conventional models and the Long short-term memory (LSTM) recurrent neural network from the deep learning models. The result indicates that the accuracy of classifying the pedagogical content reaches 92.52 \% using KNN model and 87.71 \% using LSTM model.

\end{abstract}


\section{Introduction}
\label{sec:Intro}
Billions of users create a large amount of data every day, which in a sense comes from various types of sources. This data is in most cases unorganized and unclassified and is presented in various formats such as text, video, audio, or images. Processing and analyzing this data is a major challenge that we face every day. The problem of unstructured and unorganized text dates back to ancient times, but Text Classification as a discipline first appeared in the early 60s, where 30 years later the interest in various spheres for it increased \cite{sebastiani2002machine}, and began to be applied in various types of domains and applications such as for movie review \cite{dos2014deep}, document classification \cite{kastrati2015general}, ecommerce \cite{ilmania2018aspect}, social media \cite{imran2020cross}, online courses \cite{kastrati2020weakly, itani2018understanding}, etc. As interest has grown more in the upcoming years, the uses start solving the problems with higher accurate results in more flexible ways. Knowledge Engineering (KE) was one of the applications of text classification in the late 80s, where the process took place by manually defining rules based on expert knowledge in terms of categorization of the document for a particular category \cite{sebastiani2002machine}. After this time, there was a great wave of use of various modern and advanced methods for text classification, which all improved this discipline and made it more interesting for scientists and researchers, more specifically the use of machine learning techniques. These techniques bring a lot of advantages, as they are now in very large numbers, where they provide solutions to almost every problem we may encounter. 

The need for education and learning dates back to ancient times, where people are constantly improving and trying to gain as much knowledge as possible. There are various sources of learning available today including various MOOC platforms such as Coursera, Khan Academy, Udemy, Udacity, edX, to name a few, and as technology has evolved it has contributed to better methods of acquiring knowledge that will facilitate this process. The data coming from these sources are in most cases in digital form, more specifically in the form of video and text lessons. The platforms that contain these lessons are called Massive Open Online Courses (MOOCs), where in addition to the video lesson, it also contains its textual representation called a transcript. Considering that the duration of a video lesson depends on several parameters, such as the category of video material, the platform on which the lesson is provided, the complexity of the topic, the number of instructors, and the group of lesson attendants. The duration of the lessons indirectly dictates how long the transcript will be, in other words how many words it can contain. The category shows the nature of the video and the topics that will be presented in it. As it is already known, that each video lesson belongs to a certain category, or in a group of categories, so does the transcript as well. Given this advantage, we can conclude the fact that text classification is becoming quite extensive as a discipline, where also its use can solve many challenging problems in every domain, and specifically in education domain. 

The aim of this paper is to investigate two classification techniques that are used to classify the pedagogical content, and the focus tends to compare conventional machine learning models with deep learning models, by selecting KNN algorithm for the first approach and LSTM architecture for the latter one.

To better indicate the idea we want to present, the paper will be divided into several sections, as follows: as part of literature review the main processes of classifying documents are explained, continuing with related work conducted so far in this area. In the experimental section, the design of the conventional machine learning models and deep learning models will be elaborated and the results for the each of the architectures will be presented using a number of evaluation techniques (recall, precision, F-Score, accuracy). The paper will be concluded with conclusion and future work.
 
\section{Text Classification Processes and Related Work}
\label{sec:lr}
Text mining or text analytics is one of the artificial intelligence techniques that uses Natural Language Processing (NLP) to transform unorganized and unstructured text into an appropriately structured format that will make it easier to process and analyze data. For businesses and other corporations, generating large amounts of data has become a daily routine. Analysis of this data help companies gain smarter and more creative insights regarding their services or products collected from a variety of sources in automated manner. But this analysis step requires processing a huge amount of data where the data needs to be prepared, and this is in most cases the cause of various problems. NLP is made up of five steps or phases, and they are Lexical Analysis, Syntax Analysis, Semantic Analysis, Pragmatics, and Discourse \cite{hapke2019natural}.

\begin{figure}[h]
    \begin{center}
        \includegraphics[width=150mm]{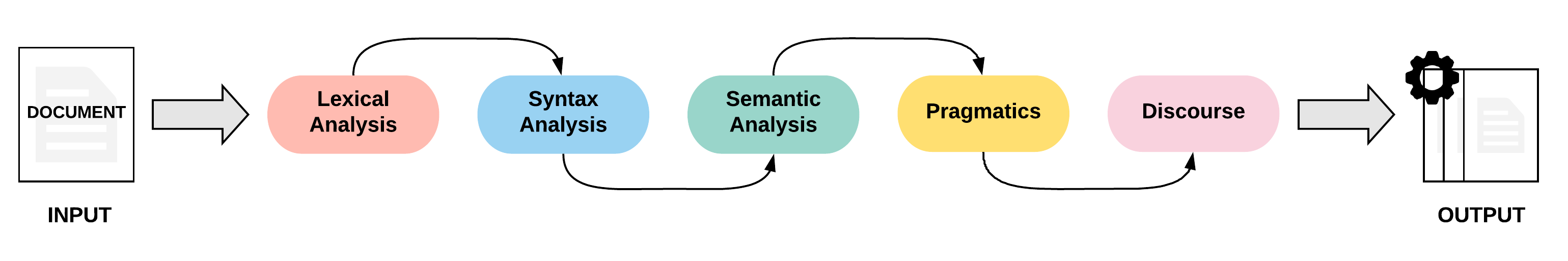}
        \caption{Natural Language Processing steps.}
        \label{fig:nlpSteps}
    \end{center}
\end{figure}

Figure \ref{fig:nlpSteps}  shows the following steps within NLP, where each of these steps will be briefly described, with the intention to understand the main concepts:

\begin{enumerate}
    \item Lexical Analysis - involves identifying the structure of a sentence, to separate words from the text, and create individual words, sentences, or paragraphs, which also includes separating punctuation from words
    \item Syntax Analysis - involves parsing words and arranging words in a sentence to have a certain meaning and relationship between them, where it is based exclusively on grammar.
    \item Semantic Analysis - implicates to analyze the grammatical structure of a word and seeks for a specific meaning in that word. The semantic analysis makes it possible to understand the relationship between lexical items.
    \item Pragmatics - means how the interpretation of a sentence is affected in its use in different situations to understand what it means and encompasses.
    \item Discourse - points out that the current sentence may depend on the previous sentence, where it can also affect the meaning of the sentence that comes after it.
\end{enumerate}

\begin{figure}[h]
    \begin{center}
        \includegraphics[width=130mm]{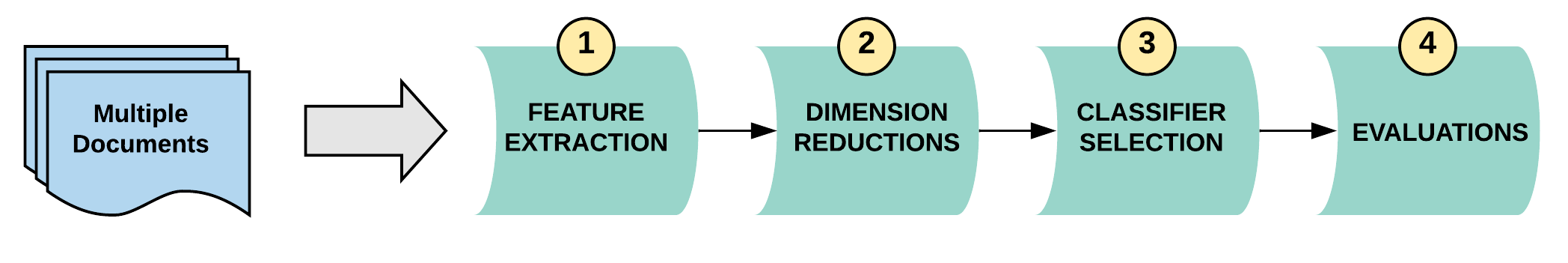}
        \caption{Four-phase model of a text classification system.}
        \label{fig:Four-phase}
    \end{center}
\end{figure}

So, the goal of text classification or text analysis is to structure and classify data to facilitate the analysis process. Today, as shown in Figure \ref{fig:Four-phase}, in order to perform text classification in the existing data, we follow the four phases emphasized by \cite{kowsari2019text}:
\begin{enumerate}
    \item Feature Extraction
    \item Dimension Reductions
    \item Classifier Selection
    \item Evaluation
\end{enumerate}

\subsection{Feature Extraction}
\label{sec:fe}

As shown in Figure \ref{fig:Four-phase}, with feature extraction as an initial phase one piece of text or document is converted into a so-called structured feature space, which will be useful to us when using a classifier. But prior to this, needs to perform data cleaning, taking care of missing data, removal of unnecessary characters or letters, in order to bring the data in an appropriate shape for extracting the features, otherwise omitting the data cleaning can directly affect negatively the performance and the accuracy of the final results. 

\begin{figure}[h]
    \begin{center}
        \includegraphics[width=130mm]{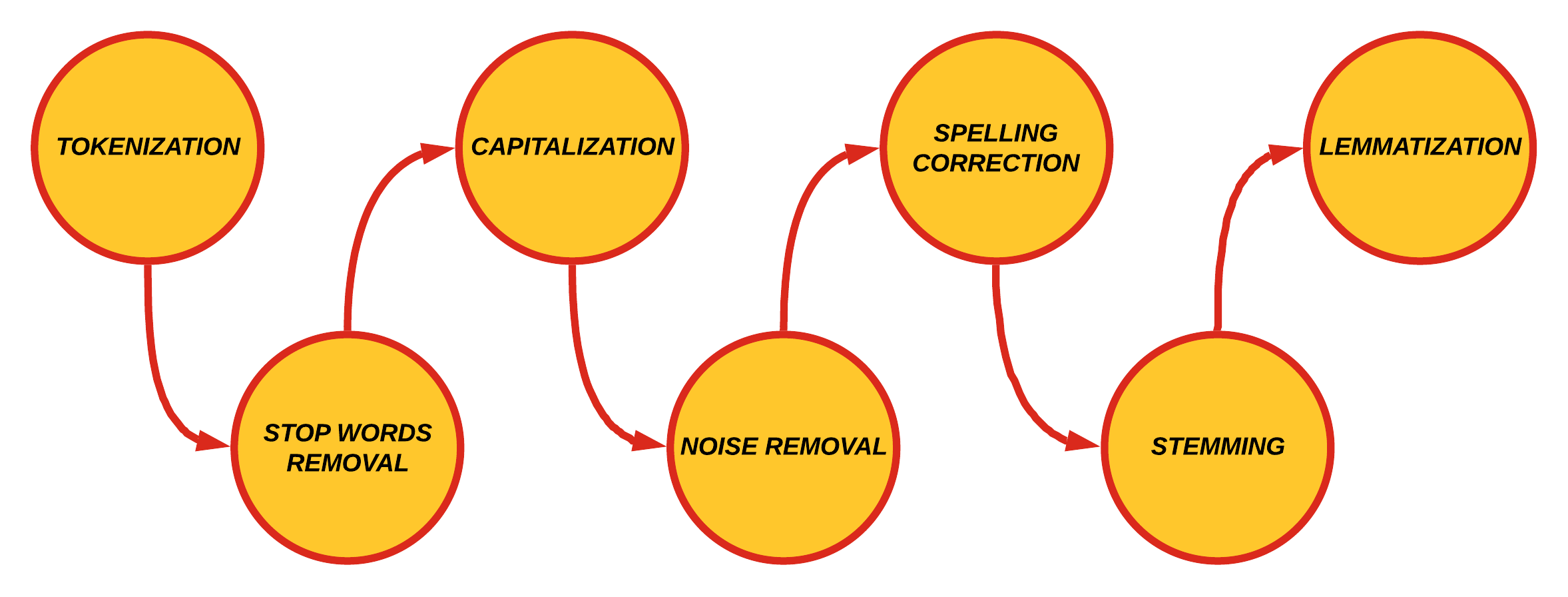}
        \caption{Techniques of data preprocessing phase.}
        \label{fig:fig3}
    \end{center}
\end{figure}

Emphasizing the importance of pre-processing data, in Figure \ref{fig:fig3}, are depicted a number of processes that are followed to clear the data and prepare it for further processing \cite{kowsari2019text}. Such processes as:

\begin{itemize}
    \item Tokenization - is the process of separating a piece of text into smaller units called tokens. The way the token is formed is based on a delimiter, which in most cases is space. Also, tokens can be words or sub-words, but also at a lower level, based on characters.
    \item Stop Words - are words that are commonly used in one language, that are unnecessary in the data processing part, and in most cases are ignored because they take up more space in the database, and affect longer processing times. In English stop words are words like: "a", "the", "an", "it", "in", "because", "what", to name a few.
    \item  Capitalization - is the part where it is necessary to identify the correct capitalization of the word, where the first word in the sentence will be automatically capitalized first. 
    \item  Noise Removal - is the process of removing characters, numbers, and parts of text that affect your analysis. These characters can be some special characters, punctuation, source code removal, HTML code removal, unique characters that represent a particular word, numbers, and many other identifiers.
    \item  Spelling Correction - is a problem where the meaning of a particular word can be mispronounced, where the word loses its meaning. This problem can be solved in two ways: with edit distance and another with overlap using k-gram.
    \item  Stemming - is a process where more morphological variants are produced than the base word or the so-called root word. For example different morphological variants of root words "like" such as "likes", "liked", "liking" and "likely".
    \item  Lemmatization - in this technique words are replaced with root words or words that have a similar meaning, and such words are called lemmas.
    \item  Syntactic Word Representation (such as N-Gram) - is a contiguous sequence of n items from one part of the text.
    \item Syntactic N-Gram - are n-grams that are constructed using paths in syntactic trees.
        \begin{itemize}
            \item Weighted Words (such as TF and TF*IDF)
             \item  Word Embedding (such as Word2Vec, GloVe, FastText)
        \end{itemize}
\end{itemize}

\subsection{Dimension Reductions}
As we can conclude from the name itself that in this step
the goal is to transform from a high-dimensional space to a low-dimensional space. The reason for this is that we strive to improve performance, speedup time, and
reduce memory complexity. There are a number of algorithms or techniques in
this step that could be implemented, such as: (i) Principal Component Analysis (PCA), (ii) Non-negative Matrix Factorization (NMF), (iii) Linear Discriminant Analysis (LDA) and (iv) Kernel PCA.

\begin{figure}[h]
    \begin{center}
        \includegraphics[width=80mm]{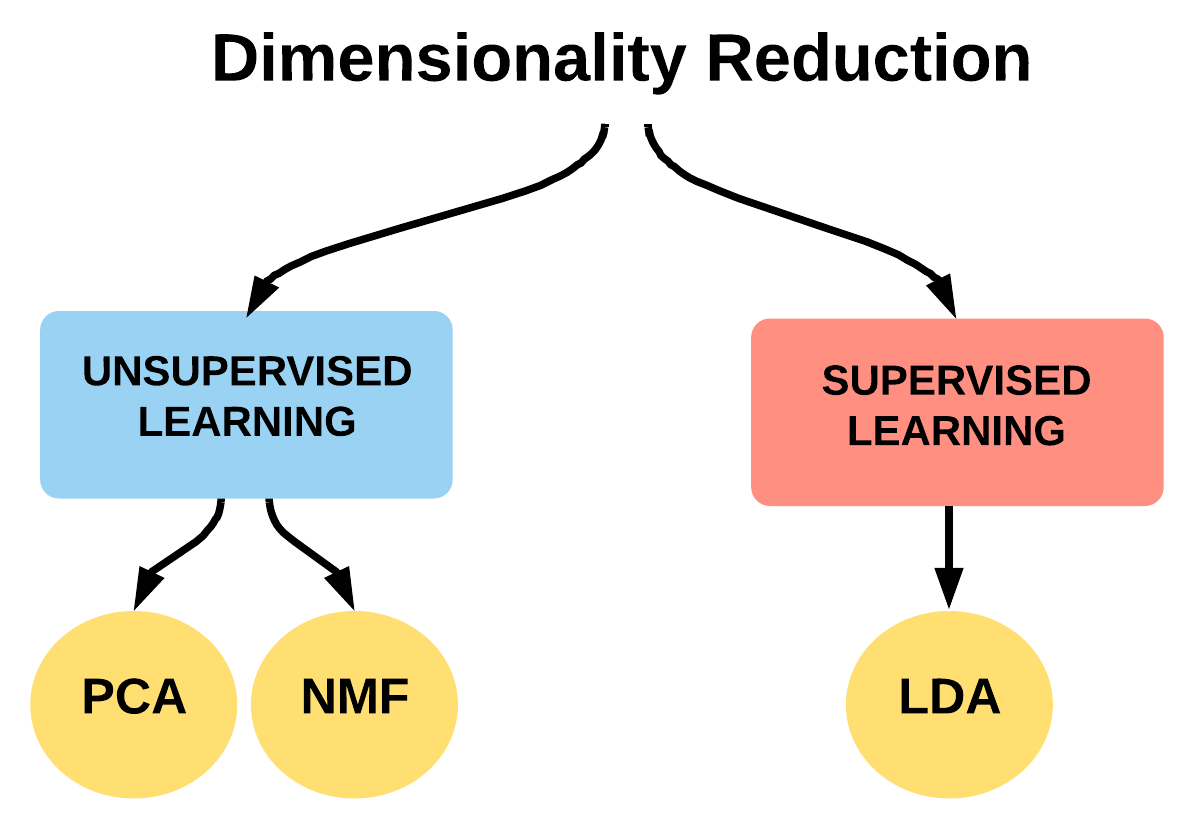}
        \caption{Categorization of dimension reductions algorithms.}
        \label{fig:dimensionalityReduction}
    \end{center}
\end{figure}

\subsection{Classifier Selection}

One of the main concerns is to choose the right classifier model that will be able to perform with a certain set of data to achieve the desired results. Choosing the right classifier model is not an easy task, and is a challenge that is also referred to in the literature as the Algorithm Selection Problem (ASP).
Every day we come across applications that use classification algorithms in some
hands. The results of the task depend on choosing the right algorithm that will
complete a particular job while showing very good performance and problem
optimization. In general, there is no single algorithm that can work for every type of
problem, and that can learn all the tasks while still being efficient, and this
phenomenon is also known as performance complementary \cite{khan2020literature}. Many factors
affect the performance of a particular algorithm, some of which is the amount of
data assigned to it for testing and training, the operating system to be executed, the
specifications of the machine on which the algorithm will be performed, and many other factors that directly or indirectly affect the selection of the algorithm. 

Some of the algorithms used for text classification are: Logistic Regression,  Naive Bayes, K-Nearest Neighbor (KNN), Support Vector Machines (SVM), Decision Trees, Random Forests, Neural Network algorithms (such as DNN, CNN, RNN) and Combination Techniques.

In our experiment we have used K-Nearest Neighbor (KNN) from the conventional models whereas LSTM recurrent neural network from the deep learning models.

\subsection{Evaluation}
One of the most important steps when creating a model for text classification is
the evaluation phase. In this phase, algorithms are analyzed or scored to assess how
efficiently they performed. 
 It should also be suggested
that comparing different parameters or metrics with this method is not an easy task.

\begin{figure}[h]
    \begin{center}
        \includegraphics[width=80mm]{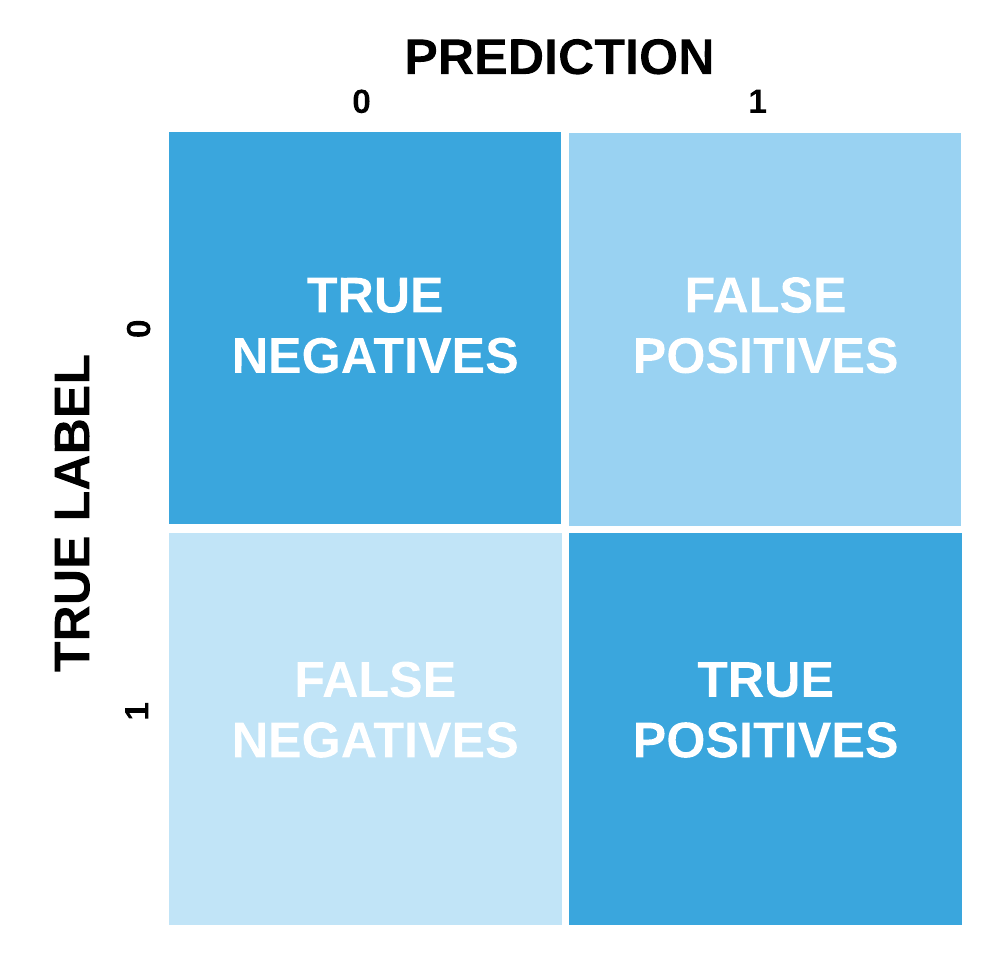}
        \caption{Confusion Matrix}
        \label{fig:fig4}
    \end{center}
\end{figure}

There is a so-called confusion matrix table (see Figure \ref{fig:fig4}) in which classification metrics such as
True Positives (TPs), False Positives (FPs), False Negatives (FNs) and True
Negatives (TNs) are calculated and presented \cite{lever2016erratum}.

Figure \ref{fig:fig4} shows a confusion matrix table in which the prediction results are
displayed horizontally, while a label that is positive or negative is shown vertically. Another evaluation technique that is lately being used is also F-Score. In this paper, in order to evaluate the experimental model the precision, recall, F-Score and Accuracy are used. 

\subsection{Related Work}
\label{sec:relatedwork}
The various technologies available today have drastically improved the way people try to gain new knowledge. Technology has greatly influenced the improvement of this process, and at the same time contributed to the development of systems that enable a more efficient and easier learning process. With this fact the use of various Massive Open Online Courses (MOOCs) begins to increase, which bring with them various opportunities, but also challenges.  Attempts to identify and analyze the opportunities and challenges of MOOCs both from pedagogical and business standpoint have led to understand how some of the very well known and successful platforms like Coursera, edX and Udacity have contributed to the improvement of their business model through various aspects,using the models for: certification model, freemium model, advertising model, job-matching model, and subcontractor model \cite{dalipi2016towards}. During the analysis of these platforms, the authors in \cite{kowsari2019text} concluded that quite a low number of students actually take assessment exams at the end of a MOOC which makes it difficult to assess whether students joining a MOOC are actually learning the content, and hence whether the MOOC is achieving its goal. One of the main components of these platforms is Learning Objects (LOs). Various techniques regarding Learning Objects (LOs) representation are presented, in which it contains pedagogical values \cite{pireva2015user}.  Using the representation features of Learning Objects will provide possibilities to personalize and customizable contents when presenting to learners along with the ability to choose an individual learning path that best suits them, aiming to maximize the learning outcome as claimed in \cite{pireva2015user}. 
There are plenty of examples where K-Means, Decision Trees, Deep Neural Network (DNN) and other machine learning techniques have been used for classification purposes \cite{dalipi2018mooc}. 
As eLearning platforms are becoming more accessible, where their main goal is to provide a smarter way of learning. The new paradigm of e-Learning also known as Cloud eLearning aiming to offer personalised learning using Cloud resources, where the main challenge is the process of content classification and matching it with learners preferences. As part of this work, the author \cite{pireva2017recommender} integrated as middle layer the recommendation systems using hierarchical clustering technique to recommend learners courses or materials that are similar to their needs before proposing a learning path using artificial intelligent automated planner. Also, paper \cite{imran2011blackboard} contributes to the classification systems in pedagogical content, with the main focus on the content classification of video lectures. The authors recommended model for the visual content classification system (VCCS) for multimedia lecture videos is to classify the content displayed on the blackboard. Through this recommended model, the authors showed over several stages how lecture videos are processed and then with a combination of support vector machines (SVM) and optical character recognition (OCR) classifies visual content, text and equations \cite{imran2011blackboard}. Furthermore in \cite{imran2016pedagogical}, researchers presented the classification and organization of pedagogical documents using domain ontology. 

In one of the previous studies \cite{chatbri2017automatic}, the authors of this paper presented a technique for automatic classification of MOOC videos, where the first step is to extract transcripts from video and then convert them into image representation using a statistical co-occurrence transform. After that, a CNN model with a dataset was implemented which was collected from Khan Academy with a total of 2545 videos, in order to evaluate the technique presented in the paper. Based on label accuracy, the best results were achieved with the CNN model, with the value of 97.87\%. Also, similar work has been carried out by Imran, Kurti and Kastrati in \cite{kastrati2019integrating} where they have proposed a video classification framework, consisting of three main modules: pre-processing, transcript representation, and classifier. In this paper, it was concluded that much better classification results were achieved with general-level than with specific-level, argued with the fact of class overlap that the specific-level category contains.

This paper aims to classify the pedagogical content using two different algorithms, K-Nearest Neighbour as an conventional machine learning model and Long short-term memory (LSTM) as an artificial recurrent neural network architecture used in deep learning. 

\section{Methodology}
\label{sec:experiment}

In this section is given the methodology used during the research and the experimental part. Initially a brief introduction regarding the dataset is given, and continuing with explanation of the architectures that are modelled to classify pedagogical content. 

Python technology is used for the whole experiment, and specifically to implement the KNN model is used the built-in functions and modules of scikit-learn library, whereas for the implementation of the RNN model is used Keras library, that runs on top of Tensorflow. In the following subsections, the used dataset as part of this experiment is described in detail, following with both models, the KNN and LSTM.

\subsection{Dataset}

The process of collecting and reviewing data is not an easy task, and in most cases requires a lot of research and finding relevant data that are used to achieve the desired results. The dataset \cite{kastrati2020wet} used in this paper for the experimental purposes is used in \cite{kastrati2019integrating} and it is modelled from scratch. This dataset consists of a total of 12,032 videos collected from the Coursera platform from more than 200 different courses. Coursera categorizes courses into a 2-level hierarchical structure from general level to fine-grained level. The general level consists of 8 categories, the specific level of 40 categories, and the course level of a total of 200 categories. In addition to these three levels that made up the course, a video lesson transcript was also included. 

\begin{figure}[h]
    \begin{center}
        \includegraphics[width=150mm]{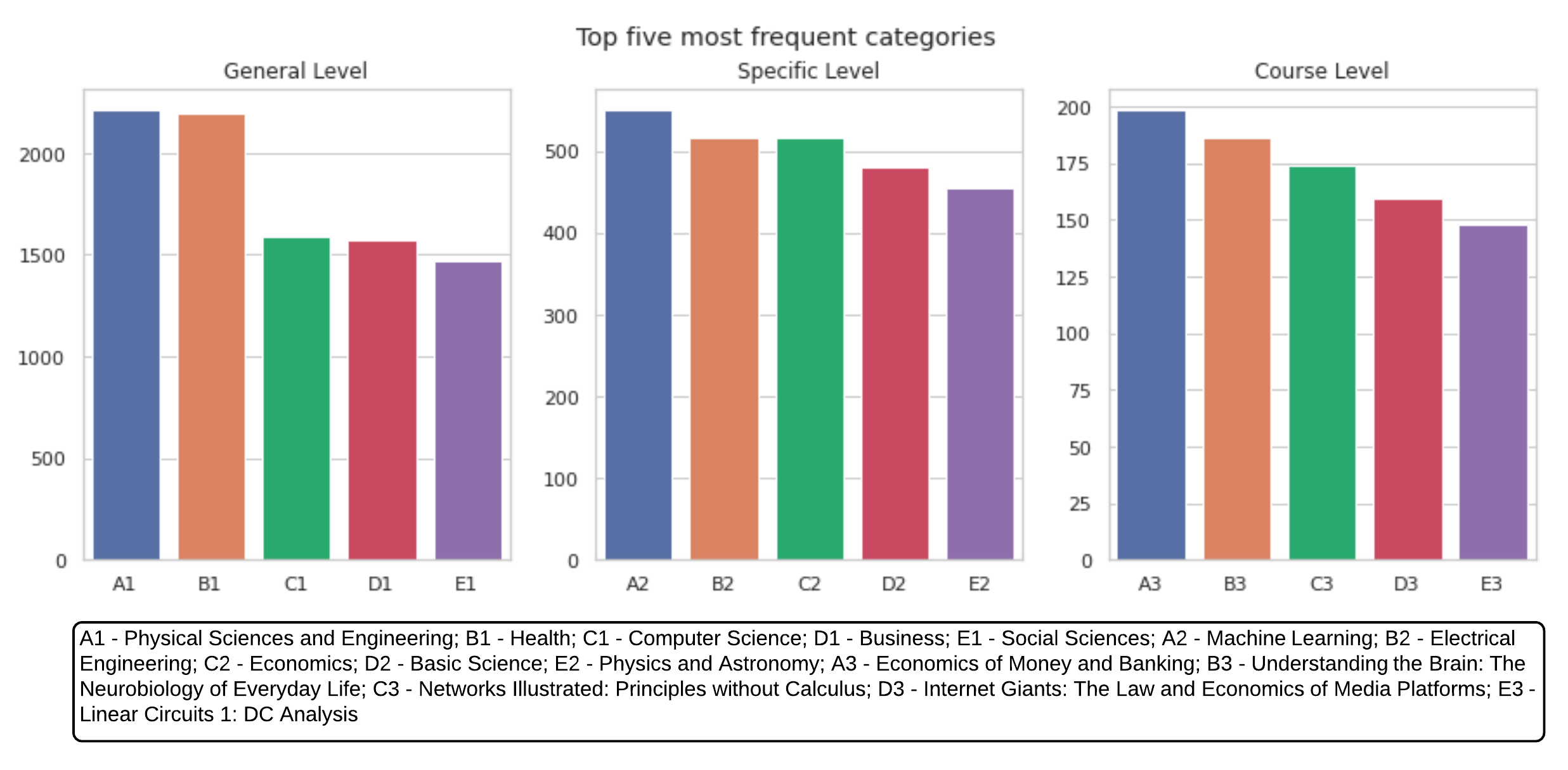}
        \caption{Top five most frequent categories for all three levels.}
        \label{fig:top5-new}
    \end{center}
\end{figure}

\begin{figure}[h]
    \begin{center}
        \includegraphics[width=150mm]{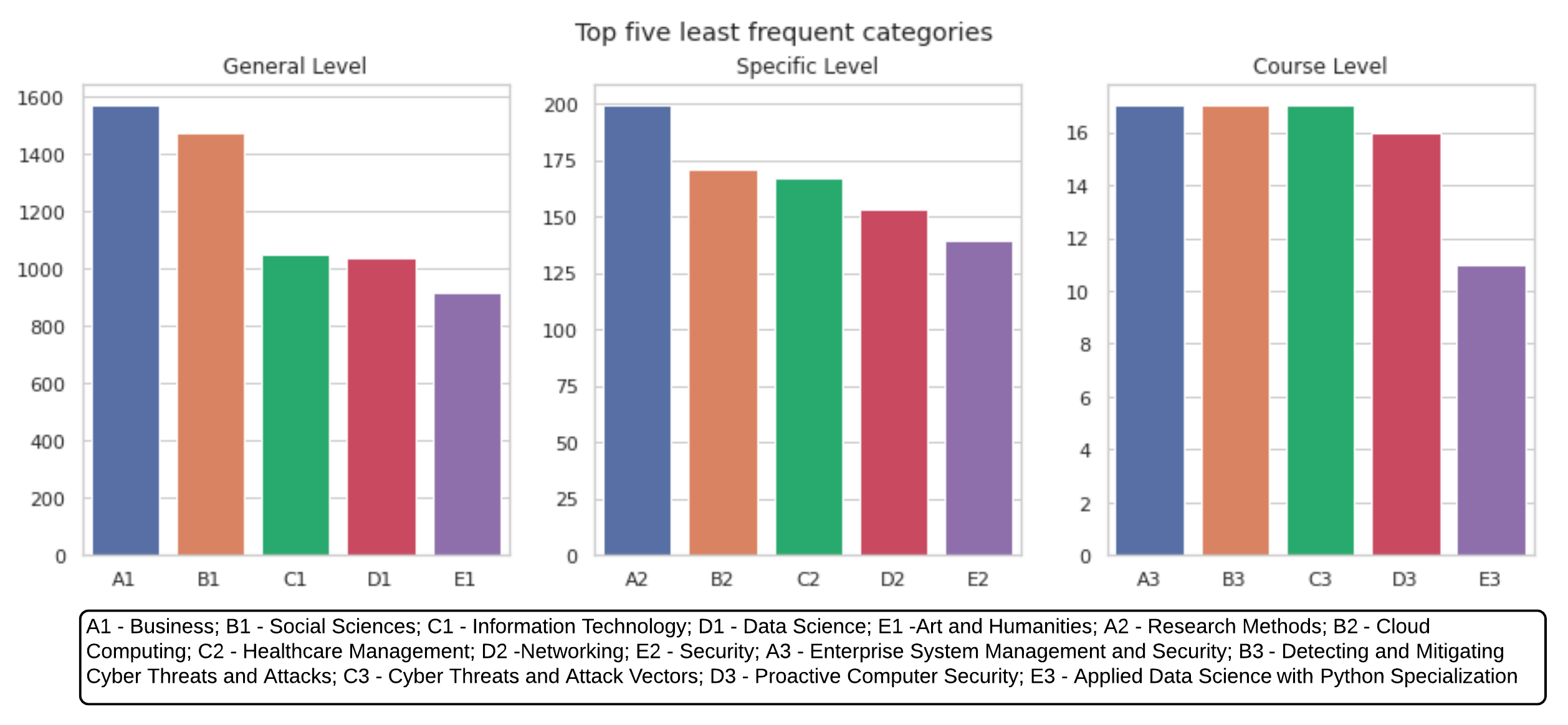}
        \caption{Top five least frequent categories for all three levels.}
        \label{fig:least5-new}
    \end{center}
\end{figure}

Figure \ref{fig:top5-new} presents the top five most frequent categories, while Figure \ref{fig:least5-new} presents the top five least frequent categories by the number of transcripts that these categories contained. 
In order for the data to be in the correct format for further analysis and modeling process, the data needs to go under pre-process phase, by preparing, cleaning, and transformed in a desired shape. The data preparation and preprocessing part depends on the given dataset, and in our case the first step after the review is to remove the noisy data (such as '[MUSIC]' which are recorded very frequently in all transcript records). Following the steps depiced in Figure \ref{fig:fig3} the entire textual content of the transcript is converted into lowercase, and removed the non-letters characters. Further, the stopwords are removed from the transcript where it helped us reduce the derived words to their particular word stem or root word as explained in \ref{sec:fe}. The dataset is transformed finally into the desired shape after finishing the lemmatization process, and it is ready to be used for both architectures that we have modelled, KNN and LSTM described further in the following subsections. 

\subsection{K-Nearest Neighbour model}

\textit{K-Nearest Neighbors (KNN)} is one of the techniques that is used in both classification and regression. It is known that KNN has no model other than collecting the entire dataset, and there is no need for learning. The predictions made with the KNN for the new data point are by searching the entire dataset for the K most similar instance (so-called neighbors) in relation to the output variant of the K instance \cite{brownlee2016master}.

There are a number of steps that the KNN algorithm goes through, such as:

\begin{enumerate}
    \item Modify K with the number of specific neighbors.
    \item Calculate the distance between the available raw data examples.
    \item Sort the calculated distances.
    \item Get the labels of top K entries.
    \item Generated prediction results for the test case.
\end{enumerate}

In this experiment, while implementing the KNN model, immediately after the process of cleaning and preparing data, is built a dictionary of features, which transforms documents to feature vectors and convert the transcripts of documents to a matrix of token counts using CountVectorizer method. Then, the count matrix is transformed to a normalized tf-idf representation using TfidfTransformer method. After this is identified the exact number of neighbors which in our case resulted in 7 neighbors. To train the classifier, the dataset is divided into two subsets:  80\% for training and 20\% for testing. Where the latter subset is used to predict the category for each input text record.

\begin{figure}[h]
    \begin{center}
        \includegraphics[width=165mm]{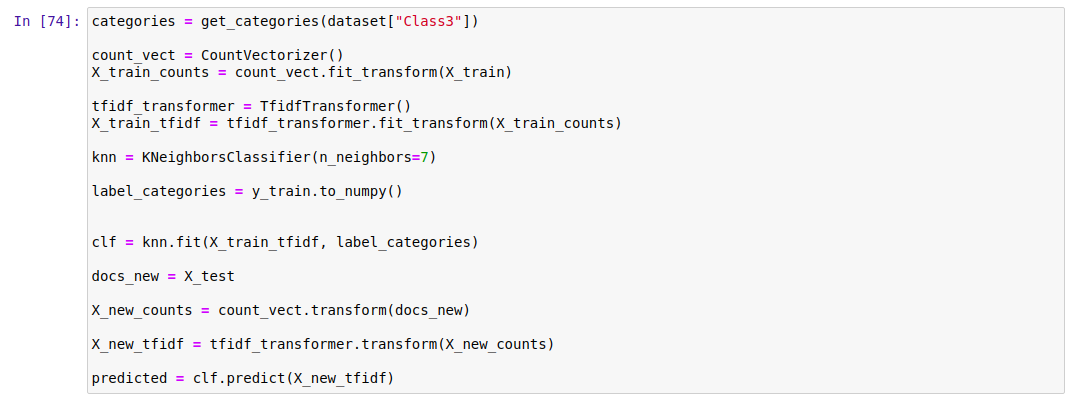}
        \caption{Implementation of KNN Classifier.}
        \label{fig:knnCode}
    \end{center}
\end{figure}

Figure \ref{fig:knnCode} shows a screenshot of implementation of our KNN classifier using Python technology and \textit{scikit-learn} library, as mentioned in Section ~\ref{sec:experiment}.

\subsection{Long short-term memory model}

\textit{Recurrent Neural Networks (RNN)} are types of artificial neural networks that allow previous outputs to be used as inputs while having hidden states \cite{amidi2018vip}. These algorithms are mostly used in fields such as: Natural Language Processing (NLP), Speech Recognition, Robot Control, Machine Translation, Music Composition, Grammar Learning, and many others. Typically, a feedforward network maps one input to one output. But as such, the inputs and outputs of neural networks can vary in the length and type of networks used for different examples and applications \cite{medsker1999recurrent}.

\begin{figure}[h]
    \begin{center}
        \includegraphics[width=165mm]{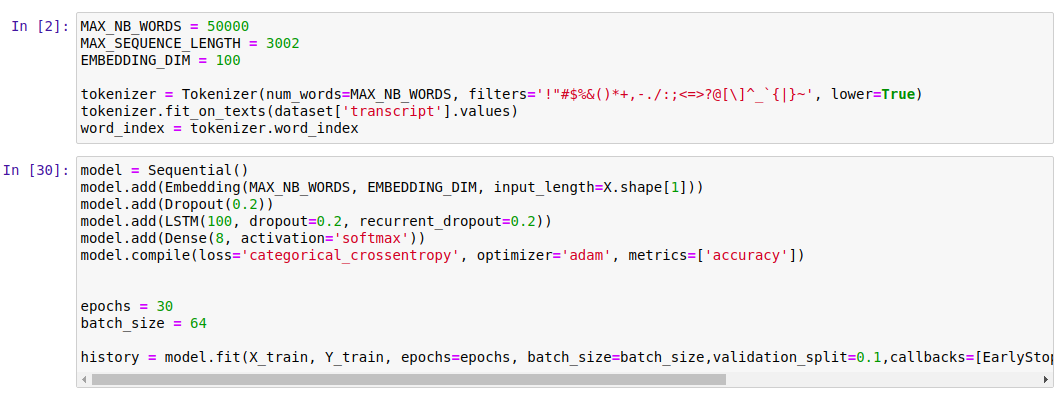}
        \caption{Implementation of LSTM model.}
        \label{fig:lstmCode}
    \end{center}
\end{figure}

Figure \ref{fig:lstmCode} shows the implementation of our LSTM model, where in this experiment in order to implement the RNN model, we used the LSTM architecture that remembers values over arbitrary intervals. As part of this architecture firstly are created Sequence models as the input layer to our network, then adding the Embedding layer which encodes to integer values the textual data entered as input, and as a result of this layer each word is then represented by a unique integer. 

For this layer, we have specified three required parameters with their respected values:

\begin{itemize}
    \item Maximum number of words - which in our case is 50000.
    \item Embedding Dim - 100.
    \item Input length - shape of X value which for us is 3002.
\end{itemize}

Further are dropped out hidden and visible units between the layers in the network, with a dropout rate of 0.2, the same value is for recurrent dropout as well. 
This is followed by the implementation of LSTM layer, and Dense layer to which we passed as the first parameter the number of units denoting the dimensionality of the output space, which in our case depends on the number of categories that are selected to classify, and as the second parameter the activation function, in this case is chosen the softmax function. And as a final step, is used categorical\_crossentropy as a loss function, and Adam as an optimizer of the network. To prevent underfitting or overfitting of the network, and to select the appropriate number of training epochs is used EarlyStopping with 'val\_loss' as a monitoring metric with patience of 3 epochs.

\section{Results}
\label{sec:results}

Table \ref{table:1} shows the classification results with the conventional model using K-Nearest Neighbours algorithm. As shown in Table \ref{table:1}, the \textit{general level} based on the precision metric has shown a very good result, 92.63\% of accuracy whereas 87.89\% accuracy is estimated by precision metrics \textit{specific level}. And at the \textit{course level}, also based on the precision metric reaches 78.59\%.
Analyzing the results for all three levels, we notice that the percentage of accuracy decreases from the upper level (general level) up to the lower level (course level). In our case, the general level consists of 8 sub-categories, the specific level of 40 sub-categories, and the course level consists of 200 sub-categories. From this we can infer that that the number of sub-categories for a single level by which the video is classified on the Coursera platform differs in each level.

\setlength{\tabcolsep}{20pt}
\renewcommand{\arraystretch}{1.5}
\setlength{\arrayrulewidth}{0.2mm}
\begin{table}[h]
\caption{Classification results with K-Nearest Neighbours.}
\begin{tabular}{|c|c|c|c|c|}
\hline
Category       & Precision (\%) & Recall (\%) & F1 Score (\%) & Accuracy (\%) \\ \hline
General Level  & 92.63          & 92.52       & 92.53         & 92.52         \\ \hline
Specific Level & 87.89          & 87.58       & 87.49         & 87.58         \\ \hline
Course Level   & 78.59          & 76.73       & 76.11         & 76.73         \\ \hline
\end{tabular}
\label{table:1}
\end{table}

Table \ref{table:2} shows the classification results with the Recurrent Neural Networks, more specifically with an Long Short-Term Memory (LSTM) architecture. Using LSTM classifier, the \textit{general level} based on the precision metric reaches 88.22\% of accuracy whereas in the \textit{specific level}, 72.31\% . Finally, at the \textit{course level}, the results shows 59.49\% of accuracy. Analyzing the results using LSTM architecture the highest accuracy is achieved at the general level, followed by a specific level, while the lowest accuracy is achieved at the course level.

\setlength{\tabcolsep}{20pt}
\renewcommand{\arraystretch}{1.5}
\setlength{\arrayrulewidth}{0.2mm}
\begin{table}[h]
\caption{Classification results with Recurrent Neural Networks}
\begin{tabular}{|c|c|c|c|c|}
\hline
Category       & Precision (\%) & Recall (\%) & F1 Score (\%) & Accuracy (\%) \\ \hline
General Level  & 88.22          & 87.71       & 87.68         & 87.71         \\ \hline
Specific Level & 72.31          & 69.93       & 70.13         & 69.93         \\ \hline
Course Level   & 59.49          & 52.91       & 53.99         & 52.91         \\ \hline
\end{tabular}
\label{table:2}
\end{table}

\section{Conclusion and future work}
\label{sec:conclusion}

In this paper are presented and discussed the classification results of the conducted experiment for all three category levels (General, Specific and Course level) using both architectures, KNN and LSTM. We can conclude that better results are achieved for levels with a smaller number of categories than for levels with a larger number of categories. In our case, as the category number increased in classes the results decreased. With this, we claim that the classification results are directly affected by the number of categories that each level contains. From results shown in Table \ref{table:1} and Table \ref{table:2} KNN reached 92.52\% of accuracy compared to LSTM with 87.71\% at general level, 87.58\%  compared to 69.93\% at specific level and  finally 76.73\% compared to 52.91\% at course level. The conducted results could be affected from several factors. First, the quantity of data required for LSTM, since a large number of categories increases the complexity of the problem, and thus requires more data to train the model. The result could have been affected due to the high similarity of different transcripts. Many of the transcripts belonged to different classes at the course level, and they had many similarities in the context of the sentences and keywords, so the model could not properly distinguish in which class the transcripts belonged. 
However, the final results gives us a spark for future work to investigate more on recurrent neural networks like, applying hyperparameters tuning, or even expand the number of architectures to further investigate the pedagogical content classification.

\bibliographystyle{unsrt}  
\bibliography{ref}  



\end{document}